\documentclass[sigconf]{acmart}
\AtBeginDocument{%
  \providecommand\BibTeX{{%
    \normalfont B\kern-0.5em{\scshape i\kern-0.25em b}\kern-0.8em\TeX}}}
\usepackage{graphicx}
\usepackage{subfigure}
\usepackage{multirow}
\usepackage{float}
\usepackage{algorithm}
\usepackage{algorithmicx}
\usepackage{algpseudocode}
\usepackage{enumitem}
\setlist[itemize]{leftmargin=*}
\setlist[enumerate]{leftmargin=*}
\setcopyright{rightsretained}
\copyrightyear{2025}
\acmYear{2025}

\acmConference[KDD '25] {Proceedings of the 1st Workshop on "AI for Supply Chain: Today and Future" @ 31st ACM SIGKDD Conference on Knowledge Discovery and Data Mining V.2}{August 3, 2025}{Toronto, ON, Canada.}

\acmBooktitle{Proceedings of the 1st Workshop on "AI for Supply Chain: Today and Future" @ 31st ACM SIGKDD Conference on Knowledge Discovery and Data Mining V.2 (KDD '25), August 3, 2025, Toronto, ON, Canada}

\acmISBN{979-8-4007-1454-2/25/08}

\acmDOI{10.1145/XXXXXX.XXXXXX}

\begin{document}


\title{Foundation Models for Demand Forecasting via Dual-Strategy Ensembling}


\author{Wei Yang}
\affiliation{%
  \institution{University of Southern California}
  \city{Los Angeles}
  \country{United States of America}}
\email{wyang930@usc.edu}

\author{Defu Cao}
\affiliation{%
  \institution{University of Southern California}
  \city{Los Angeles}
  \country{United States of America}}
\email{defucao@usc.edu}

\author{Yan Liu}
\affiliation{%
  \institution{University of Southern California; AWS Supply Chain}
  \country{United States of America}}
\email{yanliu.cs@usc.edu}



\begin{abstract}
Accurate demand forecasting is critical for supply chain optimization, yet remains difficult in practice due to hierarchical complexity, domain shifts, and evolving external factors. While recent foundation models offer strong potential for time series forecasting, they often suffer from architectural rigidity and limited robustness under distributional change. In this paper, we propose a unified ensemble framework that enhances the performance of foundation models for sales forecasting in real-world supply chains. Our method combines two complementary strategies: (1) \textbf{Hierarchical Ensemble (HE)}, which partitions training and inference by semantic levels (e.g., store, category, department) to capture localized patterns; and (2) \textbf{Architectural Ensemble (AE)}, which integrates predictions from diverse model backbones to mitigate bias and improve stability. We conduct extensive experiments on the M5 benchmark and three external sales datasets, covering both in-domain and zero-shot forecasting. Results show that our approach consistently outperforms strong baselines, improves accuracy across hierarchical levels, and provides a simple yet effective mechanism for boosting generalization in complex forecasting environments.
\end{abstract}




\maketitle

\section{Introduction}

Accurately forecasting future sales is a fundamental task in modern supply chain management \cite{ghalehkhondabi2017overview,nia2021industry,hasan2025developing,mitra2022comparative}. It drives core decisions in procurement, inventory planning, production scheduling, and logistics. Yet, despite its importance, demand forecasting remains highly challenging in practice \cite{liu2024intelligent,swaminathan2024demand}. Real-world supply chains are complex, hierarchical, and increasingly sensitive to disruptions such as economic shocks, pandemics, and geopolitical tensions \cite{trapero2024demand,kagalwala2025predictive,zhang2024enhancing,verma2024transforming}. These disruptions frequently induce regime shifts in demand signals, rendering many forecasting models brittle or unreliable. This situation creates a critical tension. While accurate forecasts are essential for operational resilience, building models that are robust, adaptive, and generalizable across diverse scenarios remains an open research problem \cite{khlie2024leveraging,jahin2024mcdfn,oyewole2024reviewing}.

Sales forecasting plays a central role in supply chain optimization \cite{theodorou2022exploring,lainder2022forecasting,kumar2024ai}. It requires accurate forecasts across product, store and regional levels while accounting for seasonality, promotions and external disruptions. The M5 competition has emerged as a key benchmark in this domain, stimulating diverse modeling strategies \cite{hewamalage2021look,seaman2022applicability,kolassa2022commentary,makridakis2022m5}. Top solutions fall into three paradigms: tree-based models like LightGBM\cite{ke2017lightgbm}, which rely on rich features and hierarchical ensembling; neural approaches such as DeepAR\cite{salinas2020deepar}, which leverage autoregressive structures and distribution-aware objectives; and hybrid or statistical methods that remain effective under sparse or intermittent demand. Recently, foundation models like Chronos \cite{ansari2024chronos} and TEMPO \cite{cao2023tempo} have gained traction by pretraining on large-scale time series to enable zero-shot generalization. While promising in bypassing task-specific tuning, these models often underperform when faced with domain shifts or hierarchical imbalances \cite{gongtowards,ma2024survey,jin2024position}. This highlights the need for new methods that can amplify their strengths while mitigating structural brittleness, particularly in complex, real-world supply chain environments.

Despite progress in time series modeling, single-model forecasters remain limited by the trade-off between bias and variance \cite{liu2024using,munoz2024literature}. Models based on trees, neural networks or foundation architectures often rely on fixed assumptions that do not adapt well across forecasting conditions. In supply chains, these limitations are amplified. Demand varies widely across products, stores and time periods, and often shifts due to promotions, seasonality or external events \cite{hasan2025developing,mitra2022comparative,khlie2024leveraging}. Even foundation models trained on diverse data can struggle in zero-shot settings when the target domain differs from the pretraining distribution. These failures lead to blind spots across segments and unstable predictions across different hierarchy levels. To address these issues, prior work has explored combining models. Mixture-of-experts methods are one example \cite{shi2024time,vallarino2024dynamic,alkilane2024mixmamba}. They aim to select or weight models dynamically, but often depend on specialized gating functions or fixed architecture designs. As a result, they can be sensitive to noise and fail to generalize under structural shifts. In real-world supply chains, where both data distribution and domain structure vary, such methods often lack robustness \cite{li2024enhancing,oyewole2024reviewing}. This highlights the need for more flexible ensemble strategies that can integrate diverse model perspectives while remaining stable across different forecasting scenarios.


To overcome the limitations of single-model forecasting in complex supply chain environments, we develop an ensemble framework that combines hierarchical structure awareness with architectural diversity. This design is guided by two key insights. First, supply chain data exhibits inherent hierarchies across stores, categories, and regions. Modeling each semantic level separately allows better alignment with localized demand patterns and improves generalization across structural variations. Second, different forecasting architectures—such as tree-based models, recurrent networks, and transformers distinct inductive biases and capture complementary aspects of temporal dynamics. By integrating these perspectives, our approach incorporates two coordinated strategies. Hierarchical Ensemble (HE) partitions the training and inference process across semantic levels to promote subgroup specialization. Architectural Ensemble (AE) combines predictions from heterogeneous backbones to mitigate model-specific variance and improve robustness. Together, these components enable foundation models to produce more accurate and stable forecasts under both in-domain and zero-shot conditions.


Our contributions are summarized as follows:

\begin{itemize}
\item We propose a unified ensemble framework that enhances the performance of foundation models for supply chain forecasting.  The framework boosts performance by strategically orchestrating how models are used and their predictions combined, thereby leveraging their collective strengths without requiring modifications to their underlying architecture.

\item We introduce two complementary strategies: Hierarchical Ensemble (HE), which models group-specific patterns through semantic partitioning of the data, and Architectural Ensemble (AE), which integrates multiple forecasting backbones to improve robustness under distribution shifts.

\item We conduct extensive experiments on the M5 benchmark and external sales datasets, demonstrating that our approach consistently outperforms strong baselines in both in-domain and zero-shot forecasting scenarios.
\end{itemize}

\section{Related Work}
\subsection{Sales Forecasting Methods}
Sales forecasting has long served as a core task in retail supply chains \cite{hasan2025developing,mitra2022comparative,theodorou2022exploring,lainder2022forecasting,kumar2024ai}, with the M5 competition emerging as a de facto benchmark for evaluating forecasting accuracy under hierarchical and sparse conditions. Tree-based models, particularly LightGBM, have dominated the leaderboard by capturing non-linear interactions and enabling fine-grained control across store, category, and SKU levels~\cite{kaggle-top1}. These models typically rely on engineered lag features, recursive training, and Tweedie objectives to address the skewness and intermittency of sales data. In parallel, purely exogenous models eschew autoregressive signals, instead framing the task as probabilistic classification using calendar and pricing features~\cite{kaggle-top2}. This shift allows better handling of cold-start or promotion-driven scenarios, especially at the lowest levels of the hierarchy. To reduce redundancy across similar sequences, transfer learning methods share model parameters across related units~\cite{wellens2022transfer}, while hybrid frameworks dynamically combine statistical and learning-based models based on local series characteristics~\cite{kolassa2022commentary}. Neural approaches have also gained traction. Variants of DeepAR incorporate multi-step rolling forecasts, hierarchical embeddings, and Tweedie loss to model zero-heavy counts~\cite{kaggle-top3}. Distributional models such as GAMLSS improve interval calibration~\cite{ziel2022m5}, while simulation-based quantile forecasts align more directly with inventory-level decisions~\cite{spiliotis2021product}.


\subsection{Foundation Models for Time Series Forecasting}
Transformer-based models have become foundational in time series forecasting due to their scalability and ability to capture long-range dependencies~\cite{zhou2021informer,liu2022pyraformer,chen2024pathformer,zhang2023crossformer,cao2023large,niu2023time}. Early variants such as \textsc{Autoformer} and \textsc{PatchTST} introduce trend decomposition and patch-based attention, respectively, offering architectural inductive biases well-suited for complex temporal dynamics.

Recent work extends this line by exploring large language models (LLMs) for time series~\cite{liu2025timecma,ye2024domain,cao2024timedit,niu2024mixture,jia2024gpt4mts,li2025climatellm}. For example, \textsc{LLMTime} and \textsc{GPT4TS} adopt GPT-style models for prompt-based forecasting~\cite{jin2023time,zhou2023one}, enabling zero-shot generalization through autoregressive generation. \textsc{TEMPO} improves generality via trend-seasonality decomposition with soft prompts~\cite{cao2023tempo}, while \textsc{UniTime} aligns forecasts with domain-specific instructions to enhance transferability~\cite{liu2024unitime}.

Another direction focuses on training foundation models directly on time series, avoiding text pretraining. \textsc{CHRONOS}~\cite{ansari2024chronos} tokenizes quantized series to build transformer-based models from scratch, and \textsc{MOIRAI}~\cite{woo2024unified} extends this approach to multivariate and irregular sequences. Large-scale variants like \textsc{TimeGPT}, \textsc{Lag-Llama}, and \textsc{TimesFM} aim to generalize across diverse domains~\cite{garza2023timegpt,rasul2023lag,das2024decoder}.

However, these models often overlook external semantics and may struggle under distribution shifts. Our work builds on this foundation by integrating ensemble learning with foundation models to improve robustness and forecasting accuracy across real-world supply chain contexts.

\section{The Proposed Method}

\subsection{Foundation Model for Time Series Forecasting}

We begin with a powerful backbone: a pre-trained foundation model for time series forecasting. Inspired by recent advances in large language models (LLMs), foundation models for temporal data aim to learn universal representations from large-scale, multi-domain time series~\cite{cao2023tempo,ansari2024chronos,garza2023timegpt}. These models possess strong zero-shot and transfer capabilities, enabling cross-domain generalization without task-specific retraining. Specifically, we refer to a “single model” as a unified forecasting architecture that includes both the model backbone and training strategy. This definition does not depend on internal structure, such as the use of mixture-of-experts or parameter sharing, but rather treats each independently trained configuration as a distinct model. In this work, we adopt a transformer-based architecture $\mathcal{F}_\theta$, parameterized by $\theta$, pre-trained on diverse temporal datasets and fine-tuned for hierarchical sales prediction on the M5 dataset.

Given an input sequence $x_{1:T} = [x_1, x_2, \ldots, x_T]$ representing historical sales and associated covariates, the model predicts a future horizon of length $H$ via:
\begin{equation}
\hat{y}_{T+1:T+H} = \mathcal{F}_\theta(x_{1:T})    
\end{equation}

The foundation model captures both short- and long-term dependencies, and is capable of adapting to novel item-store combinations in zero-shot settings. However, despite its expressive power, it often suffers from empirically observed biases in underrepresented domains, as shown in Table~\ref{tab:main_exp} by increased WRMSSE at lower levels (Levels 10–12), motivating the need for robust ensemble strategies.

\subsection{Hierarchical and Architectural Ensemble Framework}
Our method builds on the hierarchical and architectural considerations introduced in the design. We formalize this through an ensemble framework that combines semantic partitioning and model diversity to enhance forecasting performance. The framework is composed of two components: \textbf{Hierarchical Ensemble (HE)} and \textbf{Architectural Ensemble (AE)}. The Hierarchical Ensemble targets the structural organization of supply chain data. We partition the training data along semantic dimensions such as store, category, and region, and train specialized models for each group. This encourages localized learning and captures subgroup-specific temporal patterns that global models often miss. During inference, predictions are aggregated across levels to preserve consistency while retaining fine-grained accuracy.

The Architectural Ensemble addresses the variability in inductive biases across model types. We instantiate a set of diverse forecasting backbones, including statistical, recurrent and transformer-based models. These models are trained on the same data scope and generate independent forecasts. Their outputs are combined through weighted aggregation to reduce variance and increase robustness under distributional shift. By jointly leveraging hierarchical structure and architectural diversity, the ensemble achieves greater adaptability across forecasting scenarios, including both in-domain and zero-shot settings.

\subsubsection{Hierarchical Ensemble Learning (HE)}
Retail time series data are inherently hierarchical, with different levels exhibiting distinct statistical patterns. We exploit this structure by training independent models $\{\mathcal{M}_{\ell}^{(i)}\}$ for each group $i$ within a granularity level $\ell \in \{\texttt{store}, \texttt{store+category}, \texttt{store+dept}\}$. For each level, a local model $\mathcal{M}_{\ell}^{(i)}$ is trained only on the subset of data corresponding to its group $i$ (e.g., a specific store). During inference, the prediction $\hat{y}_{\ell}^{(i)}$ for a given item is obtained from its corresponding model at level $\ell$:

\begin{equation}
\hat{y}_{\ell}^{(i)} = \mathcal{M}_{\ell}^{(i)}(x_{1:T}^{(i)})
\end{equation}

The final forecast $\hat{y}$ is then computed by aggregating predictions from all levels using a weighted average:

\begin{equation}
    \hat{y} = \sum_{\ell=1}^{L} w_\ell \cdot \hat{y}_{\ell}^{(i)}, \quad \text{with} \quad \sum_{\ell=1}^{L} w_\ell = 1
\end{equation}

This multi-resolution modeling strategy aligns with the natural structure of the data and improves generalization across disjoint partitions.

\subsubsection{Architectural Ensemble Learning (AE)}
Different model architectures exhibit distinct inductive biases and error characteristics. For instance, window-based models (e.g., LightGBM) excel at tabular feature interactions, while sequence models (e.g., DeepAR, PatchTST) better capture temporal patterns. We denote the set of $K$ diverse backbones as $\{\mathcal{B}_1, \ldots, \mathcal{B}_K\}$, each producing a prediction $\hat{y}_k$ for the same target.

To combine their strengths and reduce structure-specific bias, we aggregate their outputs via weighted fusion:

\begin{equation}
    \hat{y} = \sum_{k=1}^{K} v_k \cdot \hat{y}_k, \quad \text{with} \quad \sum_{k=1}^{K} v_k = 1
\end{equation}

This ensemble yields smoother, more stable predictions by integrating architectural diversity, as shown by the consistent WRMSSE reductions in Table~\ref{tab:backbone_el}, and reduces variance from any single backbone. When combined with level-based ensemble, our final forecast benefits from both local specialization and global robustness.

\paragraph{Unified Ensemble Objective.}
The full prediction pipeline integrates both ensemble layers—hierarchical and architectural—into a unified framework:

\begin{equation}
    \hat{y}_{\text{final}} = \sum_{k=1}^{K} v_k \cdot \left( \sum_{\ell=1}^{L} w_\ell \cdot \hat{y}_{\ell, k}^{(i)} \right)
\end{equation}
where $\hat{y}_{\ell, k}^{(i)}$ denotes the prediction from model backbone $k$ at hierarchy level $\ell$ for group $i$. Here, each prediction $\hat{y}_{\ell, k}^{(i)}$ corresponds to a particular forecasting group $i$, and the summation is applied within each group. Final aggregation is conducted per group instance. For simplicity, we assign equal normalized weights during ensembling.

This hierarchical-heterogeneous ensemble strategy serves as a structural regularizer, harmonizing local specialization with model diversity, and significantly boosts forecast accuracy across all levels.

\section{EXPERIMENTS}
In this section, we conduct extensive experiments to answer the following research questions:

\begin{itemize}
\item \textbf{RQ1:} How effective is our ensemble framework in improving forecasting accuracy on supply chain benchmarks?

\item \textbf{RQ2:} Can our method enhance the zero-shot generalization ability of foundation models when applied to unseen sales datasets with different distributions?

\item \textbf{RQ3:} Does our ensemble framework consistently improve forecasting accuracy across different datasets under full-shot training conditions?

\item \textbf{RQ4:} Why does ensemble learning improve performance, and what structural dynamics underlie its effectiveness?
\end{itemize}

\subsection{Experimental Settings}

\subsubsection{Datasets}
Our experiments are conducted on four real-world sales forecasting datasets. \textbf{M5 Forecasting}\footnote{\url{https://www.kaggle.com/competitions/m5-forecasting-accuracy/overview}} is the primary dataset used for model training and adaptation. It consists of daily sales records for 30{,}490 Walmart items across multiple hierarchical levels including state, store, category and department. The dataset includes rich covariates such as calendar events and item prices, and defines WRMSSE as the official evaluation metric to emphasize both scale and aggregation consistency.

To evaluate the generalization of foundation and ensemble models, we use three real-world sales forecasting datasets from domains beyond the training data. Specifically, (1) \textbf{Sales1 (Walmart Promo)}\footnote{\url{https://www.kaggle.com/competitions/walmart-recruiting-store-sales-forecasting/overview}} includes weekly sales data with markdown events and external economic indicators; (2) \textbf{Sales2 (Store-Item Benchmark)}\footnote{\url{https://www.kaggle.com/competitions/demand-forecasting-kernels-only/overview}} is a clean benchmark dataset containing five years of daily sales across various stores and product SKUs; and (3) \textbf{Sales3 (Balkan Retail)}\footnote{\url{https://data.4tu.nl/articles/_/14406134/1}} is a real-world monthly dataset spanning seven years, covering sales and pricing information for top-selling items across multiple business units in the Balkan region. 

\subsubsection{Evaluation Metrics}
We adopt different evaluation metrics based on the characteristics of the target datasets. For the M5 forecasting task, we follow the official competition protocol and use the \textbf{Weighted Root Mean Squared Scaled Error (WRMSSE)} as the primary metric. WRMSSE is a hierarchical-aware error measure that accounts for both the scale and the relative importance of each time series in the hierarchy. It penalizes errors more heavily at aggregate levels and rewards models that maintain consistency across disaggregated units. Formally, WRMSSE extends RMSSE by introducing level-dependent weights based on sales volume and aggregation depth, enabling a unified evaluation across twelve levels of the retail hierarchy.

For the three external sales datasets, we employ standard metrics widely used in forecasting literature: the \textbf{Mean Squared Error (MSE)} and the \textbf{Mean Absolute Error (MAE)}. These metrics capture complementary aspects of prediction performance: MSE is sensitive to large errors and highlights variance, while MAE reflects median deviation and is more robust to outliers. Together, they provide a comprehensive view of model accuracy in cross-domain generalization settings.

\subsubsection{Baselines}
We compare our method against the following representative forecasting models:
\begin{itemize}
    \item \textbf{LightGBM}~\cite{ke2017lightgbm}: A strong gradient boosting model with handcrafted temporal and categorical features, widely adopted in M5 competition solutions.

    \item \textbf{DNN}~\cite{sze2017efficient}: A feedforward neural network baseline trained on static and lag-based features, representing classical deep learning approaches without temporal modeling.

    \item \textbf{DeepAR}~\cite{salinas2020deepar}: An RNN-based probabilistic model that forecasts future values by learning autoregressive conditional distributions across multiple time series.

    \item \textbf{PatchTST}~\cite{nie2022time}: A Transformer model that segments input sequences into patches and processes them in a channel-independent fashion for efficient long-range forecasting.

    \item \textbf{TEMPO}~\cite{cao2023tempo}: A foundation model pretrained on time series data using trend-seasonality decomposition and prompt-based adaptation to capture domain-invariant temporal representations.

    \item \textbf{Chronos}~\cite{ansari2024chronos}: A pre-trained Transformer model that tokenizes time series into quantized sequences and learns probabilistic forecasts via language modeling objectives.
\end{itemize}

\subsubsection{Implementation Details}
We follow the unified experimental setup provided by~\citet{zhou2023one} to ensure consistent and reproducible comparisons across baselines.\footnote{\url{https://github.com/thuml/Time-Series-Library}} All models are implemented using PyTorch~\citep{imambi2021pytorch}, and trained on a single NVIDIA A100 GPU.

For M5 forecasting, we adopt the standard evaluation split, using the last 28 days of each series as the prediction window. We follow the WRMSSE calculation protocol as defined by the official competition, and apply scale weights across twelve hierarchical levels. For zero-shot evaluation, the foundation models are directly applied to target datasets without any fine-tuning, and we report average MSE and MAE across all prediction windows.

Model-specific configurations (e.g., input sequence length, learning rate, batch size) follow either default values from the Time-Series Library~\citep{zhou2023one} or best-practice settings reported in prior works. We perform minimal hyperparameter tuning to preserve fairness and focus our evaluation on architectural design and ensemble effectiveness.

\subsection{Effectiveness of the Ensemble Framework on M5 (RQ1)}

\subsubsection{Hierarchical Ensemble Performance}

\begin{figure}[t]
  \centering
  \includegraphics[width=0.95\linewidth]{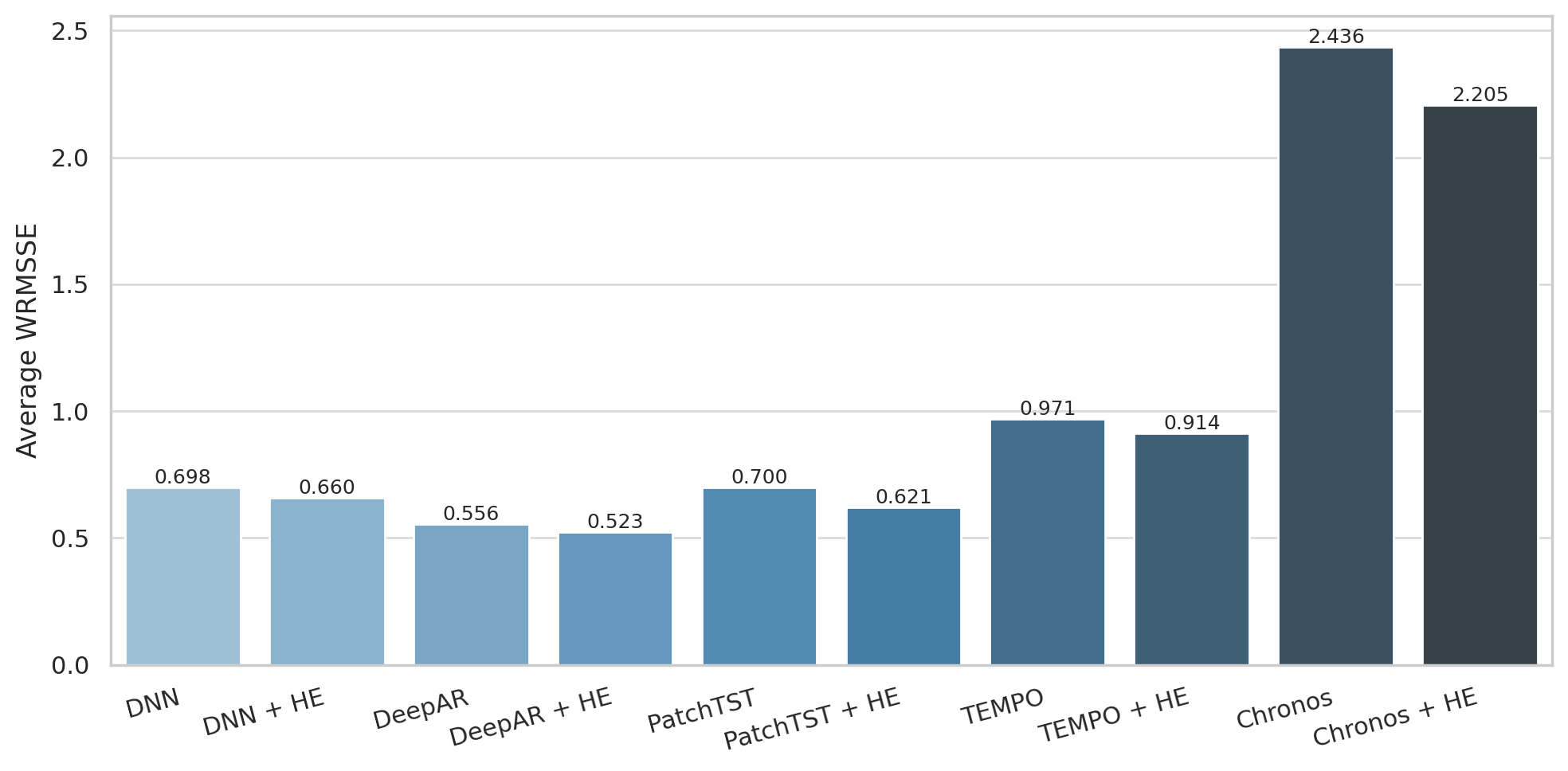}
  \caption{Average WRMSSE for backbone models and their Hierarchical Ensemble (HE) variants on the M5 dataset. HE consistently improves accuracy across model families, including tree-based, neural, and foundation models.}
  \label{fig:wrmsse_table1}
\end{figure}

Table~\ref{tab:main_exp} and Figure~\ref{fig:wrmsse_table1} reports the forecasting performance of various backbones with and without \textbf{Hierarchical Ensemble (HE)} on the M5 dataset. Results are evaluated using the WRMSSE metric across twelve hierarchical levels. Overall, applying HE consistently improves model accuracy across all architectures, demonstrating the effectiveness of hierarchy-aware specialization in capturing structured variation.

The gains are particularly notable for backbones that struggle with heterogeneity in raw form. DeepAR, for example, improves from 0.5556 to 0.5233 in WRMSSE, with especially large reductions at mid-level aggregations where local variation dominates. HE enables DeepAR to specialize across semantic groups, reducing the burden of modeling global variance. Similarly, PatchTST benefits significantly from HE, improving from 0.6997 to 0.6210. As a transformer-based model, PatchTST tends to overfit to coarse-grained signals and suffers from unstable behavior across levels. HE mitigates this by isolating distributional shifts and training within homogeneous partitions.

Even large-scale foundation models like TEMPO and Chronos show measurable gains with HE, despite being pretrained for generalization. This indicates that structural alignment remains a valuable inductive bias, even in the foundation model regime. Overall, HE serves not only as a specialization mechanism but also as a structural regularizer that improves both fine-grained fidelity and cross-level consistency.

\begin{table*}[h]
\footnotesize
\renewcommand{\arraystretch}{1.3}
\caption{WRMSSE results of different forecasting backbones with and without \textbf{Hierarchical Ensemble (HE)} applied. Results are reported across twelve aggregation levels on the M5 dataset. The Kaggle-Top1 solution corresponds to a LightGBM-based model with a hierarchical ensemble strategy. M1: Window-based, M2: RNN-based, M3: Transformer-based, M4: Foundation Model.}

\label{tab:main_exp}
\resizebox{\textwidth}{!}{
\begin{tabular}{cc|c|cccccccccccc}
\toprule
Type & Backbone & Avg. & Level 1 & Level 2 & Level 3 & Level 4 & Level 5 & Level 6 & Level 7 & Level 8 & Level 9 & Level 10 & Level 11 & Level 12 \\

\midrule
M1 & Kaggle-Top1 w/HE & \textbf{0.5230} & \textbf{0.2006} & \textbf{0.3123} & \textbf{0.3992} & \textbf{0.2843} & \textbf{0.3725} & \textbf{0.3953} & \textbf{0.4792} & \textbf{0.4815} & \textbf{0.5742} & \textbf{0.9649} & \textbf{0.9283} & \textbf{0.8838} \\
\hline
M1 & DNN & 0.6984 & 0.3405 & 0.4427 & 0.7106 & 0.4258 & 0.5741 & 0.5723 & 0.6955 & 0.8444 & 0.9187 & 0.9905 & 0.9551 & 0.9104 \\
M1 & DNN w/HE & \textbf{0.6596} & \textbf{0.2954} & \textbf{0.4183} & \textbf{0.6413} & \textbf{0.3784} & \textbf{0.5239} & \textbf{0.5379} & \textbf{0.6560} & \textbf{0.7692} & \textbf{0.8514} & \textbf{0.9862} & \textbf{0.9513} & \textbf{0.9057} \\
\midrule
M2 & DeepAR & 0.5556 & 0.2631 & 0.3469 & 0.4413 & 0.3366 & 0.4306 & 0.4312 & 0.5237 & 0.5222 & 0.6181 & 0.9560 & 0.9203 & 0.8775 \\
M2 & DeepAR w/HE & \textbf{0.5233} & \textbf{0.2054} & \textbf{0.3096} & \textbf{0.4071} & \textbf{0.2708} & \textbf{0.3815} & \textbf{0.3909} & \textbf{0.4913} & \textbf{0.4873} & \textbf{0.5920} & \textbf{0.9504} & \textbf{0.9175} & \textbf{0.8761} \\
\midrule
M3 & PatchTST & 0.6997 & 0.4626 & 0.5441 & 0.6421 & 0.5015 & 0.6032 & 0.6098 & 0.6869 & 0.6968 & 0.7652 & 0.9848 & 0.9600 & 0.9397 \\
M3 & PatchTST w/HE & \textbf{0.6210} & \textbf{0.3498} & \textbf{0.4552} & \textbf{0.5509} & \textbf{0.4028} & \textbf{0.5129} & \textbf{0.5363} & \textbf{0.6211} & \textbf{0.6238} & \textbf{0.7080} & \textbf{0.9260} & \textbf{0.8995} & \textbf{0.8654} \\
\midrule
M4 & TEMPO & 0.9706 & \textbf{0.8021} & 0.9281 & 1.1122 & \textbf{0.8197} & 0.8746 & 0.9530 & 0.9920 & 1.0953 & 1.1054 & 1.0447 & 0.9884 & 0.9321 \\
M4 & TEMPO w/HE & \textbf{0.9137} & 0.8106 & \textbf{0.8429} & \textbf{0.9088} & 0.8300 & \textbf{0.8679} & \textbf{0.8899} & \textbf{0.9354} & \textbf{0.9460} & \textbf{0.9966} & \textbf{1.0373} & \textbf{0.9788} & \textbf{0.9201} \\
\hline
M4 & Chronos & 2.4358 & 3.1928 & 2.9399 & 2.8101 & 3.1571 & 3.1303 & 2.8454 & 2.7765 & 2.5914 & 2.4350 & 1.3178 & 1.0884 & 0.9450 \\
M4 & Chronos w/HE & \textbf{2.2051} & \textbf{2.8506} & \textbf{2.6301} & \textbf{2.5219} & \textbf{2.8129} & \textbf{2.8066} & \textbf{2.5452} & \textbf{2.4973} & \textbf{2.3275} & \textbf{2.2008} & \textbf{1.2624} & \textbf{1.0651} & \textbf{0.9400} \\
\bottomrule
\end{tabular}
}
\end{table*}

\subsubsection{Architectural Ensemble Performance}

Table~\ref{tab:backbone_el} presents the results of \textbf{Architectural Ensemble (AE)}, where predictions from structurally diverse models are combined through weighted fusion. This strategy aims to mitigate the limitations of individual architectures by integrating complementary modeling perspectives. Across all configurations, AE consistently improves over single backbones, indicating that architectural diversity contributes to more stable and accurate forecasting.

The combination of LightGBM and PatchTST achieves the best overall performance, reducing the WRMSSE from 0.5230 to 0.4989. This ensemble benefits from the contrasting strengths of its components. LightGBM excels at modeling high-level aggregation with strong performance on sparse tabular data, while PatchTST adapts well to fine-grained temporal dynamics at lower levels. Their error patterns show low correlation, enabling the ensemble to cancel out systematic biases and enhance generalization.

Fusing LightGBM with DeepAR yields similar improvements, further confirming that pairing models with distinct inductive assumptions can reduce variance and improve robustness. While the PatchTST and DeepAR ensemble is slightly less effective, it still outperforms the individual models, reinforcing the value of architectural complementarity. These results support the view that AE serves as an effective bias correction mechanism. By integrating forecasts from models with divergent inductive behaviors, AE suppresses architecture-specific errors and promotes more balanced predictions across hierarchical levels. 

\begin{table*}[h]
\footnotesize
\renewcommand{\arraystretch}{1.3}
\caption{
\textbf{Architectural Ensemble (AE)} performance on the M5 dataset. Forecasts from structurally diverse backbone models, including LightGBM, DeepAR, and PatchTST, are aggregated using weighted fusion. This ensemble leverages complementary inductive biases to reduce model-specific error patterns and improve robustness across all hierarchical levels. For reference, the top-ranked Kaggle solution using only LightGBM is also included.
}
\label{tab:backbone_el}
\resizebox{\textwidth}{!}{
\begin{tabular}{l|c|cccccccccccc}
\toprule
Backbone & avg. & Level 1 & Level 2 & Level 3 & Level 4 & Level 5 & Level 6 & Level 7 & Level 8 & Level 9 & Level 10 & Level 11 & Level 12 \\
\midrule
Kaggle-Top1 & \textbf{0.5230} & \textbf{0.2006} & \textbf{0.3123} & \textbf{0.3992} & \textbf{0.2843} & \textbf{0.3725} & \textbf{0.3953} & \textbf{0.4792} & \textbf{0.4815} & \textbf{0.5742} & \textbf{0.9649} & \textbf{0.9283} & \textbf{0.8838} \\
\hline
AE (LightGBM, PatchTST) & \textbf{0.4989} & \textbf{0.1455} & 0.2907 & 0.3830 & \textbf{0.2244} & \textbf{0.3392} & 0.3688 & 0.4631 & 0.4625 & 0.5623 & \textbf{0.9528} & \textbf{0.9188} & \textbf{0.8758} \\
AE (LightGBM, DeepAR) & 0.5003 & 0.1537 & \textbf{0.2903} & \textbf{0.3792} & 0.2330 & 0.3427 & \textbf{0.3677} & \textbf{0.4627} & \textbf{0.4591} & \textbf{0.5607} & 0.9556 & 0.9212 & 0.8777 \\
AE (PatchTST, DeepAR) & 0.5216 & 0.2002 & 0.3119 & 0.4163 & 0.2601 & 0.3800 & 0.3946 & 0.4948 & 0.4954 & 0.5955 & 0.9366 & 0.9064 & 0.8674 \\
\bottomrule
\end{tabular}
}
\end{table*}

\subsection{Foundation-Model Ensemble for Zero-Shot Forecasting (RQ2)}

We evaluate the zero-shot generalization of foundation models and their ensemble-enhanced variants on real-world sales forecasting datasets outside the M5 domain. In this setting, models are applied without fine-tuning, simulating deployment to unseen markets where no target labels are available. The goal is to assess whether ensemble integration can improve robustness under distribution shift, even when foundation models are pre-trained to generalize.

Architectural Ensemble consistently improves the performance of all tested backbones. This effect is especially pronounced for PatchTST, a transformer-based model trained without cross-domain pretraining. PatchTST alone shows limited transferability and unstable error patterns across datasets. When combined with structurally distinct models through AE, its predictions become more stable and better aligned with local temporal structures, suggesting that ensemble fusion can compensate for narrow inductive biases.

Foundation models like Chronos and TEMPO exhibit stronger baseline performance, yet also benefit from AE. In particular, Chronos gains from the integration of models that specialize in coarse-grained seasonality or localized trends. For TEMPO, which already includes architectural adaptations such as decomposition and modulation, AE acts as a form of bias regularization. It smooths over residual overfitting and introduces modeling perspectives that help balance representation across varying domains. Pretraining provides broad generalization capacity, but it does not eliminate structural blind spots or ensure robustness under extreme domain shifts. AE complements this by integrating heterogeneous inductive signals, enabling more resilient zero-shot forecasting without requiring retraining or domain-specific adaptation.

\begin{table}[h]
\centering
\footnotesize
\renewcommand{\arraystretch}{1.2}
\caption{Zero-shot forecasting performance (MSE / MAE) of foundation models and their ensemble-enhanced variants across three external sales datasets. HE denotes Hierarchical Ensemble. Foundation models are pre-trained on the M5 dataset without access to target domains.}
\label{tab:zeroshot}
\resizebox{\linewidth}{!}{
\begin{tabular}{l|ccc}
\toprule
\multirow{2}{*}{\textbf{Model}} & \textbf{Sales1} & \textbf{Sales2} & \textbf{Sales3} \\
\cline{2-4}
& \textit{MSE / MAE} & \textit{MSE / MAE} & \textit{MSE / MAE} \\
\midrule
PatchTST & 7.3854 / 1.3052 & 11.7808 / 3.1509 & 2.7013 / 1.1475 \\
PatchTST w/HE & \textbf{5.4312} / \textbf{1.0393} & \textbf{8.5738} / \textbf{2.3874} & \textbf{2.4085} / \textbf{1.0712} \\
\midrule
TEMPO & 1.1690 / 0.8357 & 1.1814 / 0.8792 & 1.7293 / 0.9556 \\
TEMPO w/HE & \textbf{1.1173 / 0.8038} & \textbf{1.1503 / 0.8674} & \textbf{1.6725 / 0.9487} \\
\midrule
Chronos & 2.5253 / 1.1935 & 2.6585 / 1.1721 & 3.1250 / 1.4787 \\
Chronos w/HE & \textbf{2.3375} / \textbf{1.1372} & \textbf{2.3751} / \textbf{1.0965} & \textbf{2.8533} / \textbf{1.3514} \\
\bottomrule
\end{tabular}
}
\end{table}

\begin{table}[h]
\centering
\footnotesize
\renewcommand{\arraystretch}{1.2}
\caption{Full-shot forecasting performance (MSE / MAE) of foundation models and their ensemble-enhanced variants across three external sales datasets. HE denotes Hierarchical Ensemble. Foundation models are trained on each specific dataset.}
\label{tab:zeroshot}
\resizebox{\linewidth}{!}{
\begin{tabular}{l|ccc}
\toprule
\multirow{2}{*}{\textbf{Model}} & \textbf{Sales1} & \textbf{Sales2} & \textbf{Sales3} \\
\cline{2-4}
& \textit{MSE / MAE} & \textit{MSE / MAE} & \textit{MSE / MAE} \\
\midrule
PatchTST & 0.0420 / 0.0949 & 0.0896 / 0.2305 & 0.7161 / 0.5177 \\
PatchTST w/HE & \textbf{0.0403} / \textbf{0.0938} & \textbf{0.0881} / \textbf{0.2282} & \textbf{0.6673} / \textbf{0.4938} \\
\midrule
TEMPO & 0.0408 / 0.0939 & 0.0878 / 0.2284 & 0.5869 / 0.4751 \\
TEMPO w/HE & \textbf{0.0391 / 0.0921} & \textbf{0.0869 / 0.2263} & \textbf{0.5791 / 0.4711} \\
\midrule
Chronos & 0.0411 / 0.0933 & 0.0887 / 0.2296 & 0.6019 / 0.4838 \\
Chronos w/HE & \textbf{0.0398} / \textbf{0.0925} & \textbf{0.0875} / \textbf{0.2284} & \textbf{0.5827} / \textbf{0.4787} \\
\bottomrule
\end{tabular}
}
\end{table}

\begin{table*}[h]
\centering
\footnotesize
\caption{WRMSSE of PatchTST under different hierarchical modeling strategies on the M5 dataset. ``Store-Dept'', ``Store-Cate'', and ``Store'' denote models trained on different semantic levels. ``Ensemble'' corresponds to the Hierarchical Ensemble (HE) of the three levels.}
\label{tab:patchtst_hierarchy}
\resizebox{\linewidth}{!}{
\begin{tabular}{l|ccccccccccccc}
\toprule
Model & Avg. & Level1 & Level2 & Level3 & Level4 & Level5 & Level6 & Level7 & Level8 & Level9 & Level10 & Level11 & Level12 \\
\midrule
PatchTST (Store-Dept) & 0.6349 & 0.3374 & 0.4547 & 0.5650 & 0.3982 & 0.5069 & 0.5412 & 0.6390 & 0.6538 & 0.7571 & 0.9245 & 0.9224 & 0.9187 \\
PatchTST (Store-Cate) & 0.6387 & 0.3559 & 0.4542 & 0.5713 & 0.4118 & 0.5289 & 0.5499 & 0.6357 & 0.6563 & 0.7347 & 0.9276 & 0.9220 & 0.9159 \\
PatchTST (Store)      & 0.6339 & 0.3615 & 0.4716 & 0.5625 & 0.4173 & 0.5344 & 0.5528 & 0.6368 & 0.6330 & 0.7126 & 0.9170 & 0.9080 & 0.8994 \\
\midrule
PatchTST (Ensemble)   & \textbf{0.6210} & \textbf{0.3498} & \textbf{0.4552} & \textbf{0.5509} & \textbf{0.4028} & \textbf{0.5129} & \textbf{0.5363} & \textbf{0.6211} & \textbf{0.6238} & \textbf{0.7080} & \textbf{0.9260} & \textbf{0.8995} & \textbf{0.8654} \\
\bottomrule
\end{tabular}
}
\end{table*}

\subsection{Cross-Dataset Generalization of Hierarchical Ensemble in Full-Shot Forecasting (RQ3)}

While RQ1 establishes the effectiveness of Hierarchical Ensemble (HE) on the M5 dataset in a full-shot training setting, it remains unclear whether the same benefits generalize to other domains. In this section, we evaluate the robustness of HE across three external sales datasets by retraining foundation models from scratch on each dataset and comparing performance with and without HE.

Table~\ref{tab:zeroshot} presents the results for PatchTST, TEMPO, and Chronos when trained directly on each target dataset. Across all models and datasets, we observe consistent improvements in both MSE and MAE after applying HE. For example, PatchTST w/HE outperforms its non-ensemble variant on all datasets, with particularly notable gains on Sales3 (MSE reduced from 0.7161 to 0.6673). Similar trends are observed for TEMPO and Chronos, confirming the general effectiveness of HE even in domains outside the M5 benchmark.

These results suggest that the benefits of HE extend beyond a single dataset and are not specific to any particular domain structure. Even when retraining on diverse datasets with different temporal resolutions and demand patterns, HE consistently enhances performance. This supports the claim that HE functions as a general-purpose structural inductive bias, encouraging subgroup-level specialization and reducing variance due to global overfitting.

In contrast to RQ2, which focused on zero-shot transfer from a single source dataset (M5), this experiment demonstrates that HE retains its effectiveness in realistic full-shot deployment scenarios where training on the target domain is feasible. Together with the results from RQ1 and RQ2, this cross-dataset analysis highlights the broad applicability of the HE strategy and motivates its adoption in both transfer and direct-learning forecasting workflows.

\subsection{Why Does Ensemble Work? A Layer-Wise View (RQ4)}

To better understand the efficacy of ensemble learning, we examine the performance of PatchTST trained independently at three semantic levels: Store, Store+Department, and Store+Category and compare them with their ensemble combination. Table~\ref{tab:patchtst_hierarchy} reports the WRMSSE across all 12 evaluation levels defined in the M5 competition, which span from total sales aggregated across all stores (Level 1) to individual product-store combinations (Level 12).

A key insight emerges when aligning performance patterns with the semantic meaning of each level. Levels 1--5 correspond to coarse-grained aggregations such as total sales by state or category. At these levels, the \textit{Store+Department} model performs best, as it captures relatively stable and high-volume sales patterns associated with department-level trends. However, at mid-level aggregations (Levels 6--9), which include cross-structured dimensions such as Store+Category or State+Department, the performance advantage shifts slightly toward \textit{Store+Category}, likely due to its finer specialization along consumer preference lines. At the lowest levels (Levels 10--12), which evaluate item-level forecasts by store or region, the \textit{Store-only} model shows superior adaptability, as it has learned from more homogeneous, localized patterns.

These observations highlight that no single hierarchical partition captures all relevant signal components across levels. Each level-specific model overfits or undergeneralizes in different structural regimes. Importantly, the HE fusion of these three perspectives consistently improves or matches the best component model across all levels, resulting in an overall WRMSSE reduction from ~0.6349 to 0.6210.

From a theoretical standpoint, this reinforces the notion that structural diversity among models introduces orthogonal inductive biases, which when aggregated, cancel out local biases and yield a more balanced predictor. HE not only reduces forecast variance but acts as a structure-aware alignment mechanism, reconciling inconsistencies between different views of the data hierarchy. The ensemble is not simply averaging—it is synthesizing semantically complementary forecasts that individually dominate in specific regimes but are collectively incomplete.


\subsection{Discussion and Limitations}

Our findings reveal that effective demand forecasting in real-world supply chains increasingly hinges not on refining a single model architecture, but on orchestrating diverse inductive perspectives across structural and architectural dimensions. The dual ensemble strategies proposed in this work address complementary sources of forecasting difficulty: HE mitigates distributional fragmentation caused by semantic heterogeneity (e.g., store, category, department), while AE balances architectural biases by fusing models with distinct representational priors.

Notably, our analysis demonstrates that no single semantic partition dominates performance across all levels of the hierarchy. Instead, different granularities specialize in different structural regimes, and their ensemble synthesis yields performance superior to any individual specialization. Similarly, the combination of tree-based models and transformer-based foundation models leverages non-overlapping error characteristics, leading to more robust and stable forecasts, particularly under distribution shifts.

From a broader perspective, these results suggest that forecasting accuracy in complex domains emerges not from deeper architectures, but from structured diversity—a principle aligned with emerging trends in foundation model research. While pretrained models such as \textsc{TEMPO} and \textsc{CHRONOS} offer strong zero-shot capabilities, they remain susceptible to inductive blind spots introduced by coarse pretraining objectives or flattened hierarchies. Our ensemble framework effectively regularizes these foundation models by anchoring them to localized patterns and complementary views.

This insight carries broader implications. As supply chains become more volatile and fragmented, future forecasting systems must be modular, adaptive, and capable of reconciling signals across levels and representations. Rather than seeking universal predictors, we advocate for systems that learn to coordinate partial experts, each aligned with a semantically coherent subspace or architectural strength. Such systems not only improve accuracy but offer greater resilience, transparency, and extensibility for downstream decision-making.

Despite these advances, several limitations and open questions remain. While our ensemble strategy is effective in fusing structural and architectural diversity, it currently operates with fixed combination schemes and does not explicitly adapt to context or input uncertainty. Moreover, although foundation models pretrained on large-scale sales data demonstrate strong generalization, their robustness across domains with distinct temporal patterns or operational semantics (e.g., manufacturing or healthcare) is still not well understood. Finally, while accuracy improves significantly, the interpretability and explainability of ensemble outputs remain an open challenge, especially in decision-making environments.

\section{Conclusion}
In this paper, we present a unified ensemble framework for sales forecasting in supply chain settings, combining hierarchical structure awareness with architectural diversity to enhance the performance of pretrained foundation models. Our method integrates Hierarchical Ensemble (HE) and Architectural Ensemble (AE) to mitigate model-specific biases, capture fine-grained local patterns, and enable robust generalization across both in-domain and zero-shot scenarios. Empirical results on the M5 benchmark and three external datasets demonstrate that our approach consistently improves forecasting accuracy and stability, particularly when applied to foundation models such as TEMPO and Chronos.

Looking forward, we envision several promising research directions. First, the development of domain-specialized foundation models tailored for supply chain data could significantly enhance representational alignment, especially when incorporating structured metadata. Second, integrating LLM-based temporal reasoning offers a powerful avenue for uncovering latent causal structures and interpreting forecasting decisions. Third, Retrieval-Augmented Generation (RAG) can help forecasting models adapt to changing environments by retrieving relevant signals such as promotions, macroeconomic trends and calendar events. Finally, combining generative modeling with discriminative objectives may improve both forecast accuracy and interpretability, which is essential for decision support in supply chain operations.


\clearpage
\bibliographystyle{ACM-Reference-Format}
\bibliography{samplebase}


\begin{thebibliography}{61}


\ifx \showCODEN    \undefined \def \showCODEN     #1{\unskip}     \fi
\ifx \showDOI      \undefined \def \showDOI       #1{#1}\fi
\ifx \showISBNx    \undefined \def \showISBNx     #1{\unskip}     \fi
\ifx \showISBNxiii \undefined \def \showISBNxiii  #1{\unskip}     \fi
\ifx \showISSN     \undefined \def \showISSN      #1{\unskip}     \fi
\ifx \showLCCN     \undefined \def \showLCCN      #1{\unskip}     \fi
\ifx \shownote     \undefined \def \shownote      #1{#1}          \fi
\ifx \showarticletitle \undefined \def \showarticletitle #1{#1}   \fi
\ifx \showURL      \undefined \def \showURL       {\relax}        \fi
\providecommand\bibfield[2]{#2}
\providecommand\bibinfo[2]{#2}
\providecommand\natexlab[1]{#1}
\providecommand\showeprint[2][]{arXiv:#2}

\bibitem[\protect\citeauthoryear{Alkilane, He, and Lee}{Alkilane et~al\mbox{.}}{2024}]%
        {alkilane2024mixmamba}
\bibfield{author}{\bibinfo{person}{Khaled Alkilane}, \bibinfo{person}{Yihang He}, {and} \bibinfo{person}{Der-Horng Lee}.} \bibinfo{year}{2024}\natexlab{}.
\newblock \showarticletitle{MixMamba: Time series modeling with adaptive expertise}.
\newblock \bibinfo{journal}{\emph{Information Fusion}}  \bibinfo{volume}{112} (\bibinfo{year}{2024}), \bibinfo{pages}{102589}.
\newblock


\bibitem[\protect\citeauthoryear{Ansari, Stella, Turkmen, Zhang, Mercado, Shen, Shchur, Rangapuram, Arango, Kapoor, et~al\mbox{.}}{Ansari et~al\mbox{.}}{2024}]%
        {ansari2024chronos}
\bibfield{author}{\bibinfo{person}{Abdul~Fatir Ansari}, \bibinfo{person}{Lorenzo Stella}, \bibinfo{person}{Caner Turkmen}, \bibinfo{person}{Xiyuan Zhang}, \bibinfo{person}{Pedro Mercado}, \bibinfo{person}{Huibin Shen}, \bibinfo{person}{Oleksandr Shchur}, \bibinfo{person}{Syama~Sundar Rangapuram}, \bibinfo{person}{Sebastian~Pineda Arango}, \bibinfo{person}{Shubham Kapoor}, {et~al\mbox{.}}} \bibinfo{year}{2024}\natexlab{}.
\newblock \showarticletitle{Chronos: Learning the language of time series}.
\newblock \bibinfo{journal}{\emph{arXiv preprint arXiv:2403.07815}} (\bibinfo{year}{2024}).
\newblock


\bibitem[\protect\citeauthoryear{Cao, Jia, Arik, Pfister, Zheng, Ye, and Liu}{Cao et~al\mbox{.}}{2023a}]%
        {cao2023tempo}
\bibfield{author}{\bibinfo{person}{Defu Cao}, \bibinfo{person}{Furong Jia}, \bibinfo{person}{Sercan~O Arik}, \bibinfo{person}{Tomas Pfister}, \bibinfo{person}{Yixiang Zheng}, \bibinfo{person}{Wen Ye}, {and} \bibinfo{person}{Yan Liu}.} \bibinfo{year}{2023}\natexlab{a}.
\newblock \showarticletitle{Tempo: Prompt-based generative pre-trained transformer for time series forecasting}.
\newblock \bibinfo{journal}{\emph{arXiv preprint arXiv:2310.04948}} (\bibinfo{year}{2023}).
\newblock


\bibitem[\protect\citeauthoryear{Cao, Ye, Zhang, and Liu}{Cao et~al\mbox{.}}{2024}]%
        {cao2024timedit}
\bibfield{author}{\bibinfo{person}{Defu Cao}, \bibinfo{person}{Wen Ye}, \bibinfo{person}{Yizhou Zhang}, {and} \bibinfo{person}{Yan Liu}.} \bibinfo{year}{2024}\natexlab{}.
\newblock \showarticletitle{Timedit: General-purpose diffusion transformers for time series foundation model}.
\newblock \bibinfo{journal}{\emph{arXiv preprint arXiv:2409.02322}} (\bibinfo{year}{2024}).
\newblock


\bibitem[\protect\citeauthoryear{Cao, Zheng, Hassanzadeh, Lamba, Liu, and Liu}{Cao et~al\mbox{.}}{2023b}]%
        {cao2023large}
\bibfield{author}{\bibinfo{person}{Defu Cao}, \bibinfo{person}{Yixiang Zheng}, \bibinfo{person}{Parisa Hassanzadeh}, \bibinfo{person}{Simran Lamba}, \bibinfo{person}{Xiaomo Liu}, {and} \bibinfo{person}{Yan Liu}.} \bibinfo{year}{2023}\natexlab{b}.
\newblock \showarticletitle{Large scale financial time series forecasting with multi-faceted model}. In \bibinfo{booktitle}{\emph{Proceedings of the Fourth ACM International Conference on AI in Finance}}. \bibinfo{pages}{472--480}.
\newblock


\bibitem[\protect\citeauthoryear{Chen, Zhang, Cheng, Shu, Wang, Wen, Yang, and Guo}{Chen et~al\mbox{.}}{2024}]%
        {chen2024pathformer}
\bibfield{author}{\bibinfo{person}{Peng Chen}, \bibinfo{person}{Yingying Zhang}, \bibinfo{person}{Yunyao Cheng}, \bibinfo{person}{Yang Shu}, \bibinfo{person}{Yihang Wang}, \bibinfo{person}{Qingsong Wen}, \bibinfo{person}{Bin Yang}, {and} \bibinfo{person}{Chenjuan Guo}.} \bibinfo{year}{2024}\natexlab{}.
\newblock \showarticletitle{Pathformer: Multi-scale transformers with adaptive pathways for time series forecasting}.
\newblock \bibinfo{journal}{\emph{arXiv preprint arXiv:2402.05956}} (\bibinfo{year}{2024}).
\newblock


\bibitem[\protect\citeauthoryear{Das, Kong, Sen, and Zhou}{Das et~al\mbox{.}}{2024}]%
        {das2024decoder}
\bibfield{author}{\bibinfo{person}{Abhimanyu Das}, \bibinfo{person}{Weihao Kong}, \bibinfo{person}{Rajat Sen}, {and} \bibinfo{person}{Yichen Zhou}.} \bibinfo{year}{2024}\natexlab{}.
\newblock \showarticletitle{A decoder-only foundation model for time-series forecasting}. In \bibinfo{booktitle}{\emph{Forty-first International Conference on Machine Learning}}.
\newblock


\bibitem[\protect\citeauthoryear{Garza, Challu, and Mergenthaler-Canseco}{Garza et~al\mbox{.}}{2023}]%
        {garza2023timegpt}
\bibfield{author}{\bibinfo{person}{Azul Garza}, \bibinfo{person}{Cristian Challu}, {and} \bibinfo{person}{Max Mergenthaler-Canseco}.} \bibinfo{year}{2023}\natexlab{}.
\newblock \showarticletitle{TimeGPT-1}.
\newblock \bibinfo{journal}{\emph{arXiv preprint arXiv:2310.03589}} (\bibinfo{year}{2023}).
\newblock


\bibitem[\protect\citeauthoryear{Ghalehkhondabi, Ardjmand, Weckman, and Young}{Ghalehkhondabi et~al\mbox{.}}{2017}]%
        {ghalehkhondabi2017overview}
\bibfield{author}{\bibinfo{person}{Iman Ghalehkhondabi}, \bibinfo{person}{Ehsan Ardjmand}, \bibinfo{person}{Gary~R Weckman}, {and} \bibinfo{person}{William~A Young}.} \bibinfo{year}{2017}\natexlab{}.
\newblock \showarticletitle{An overview of energy demand forecasting methods published in 2005--2015}.
\newblock \bibinfo{journal}{\emph{Energy Systems}}  \bibinfo{volume}{8} (\bibinfo{year}{2017}), \bibinfo{pages}{411--447}.
\newblock


\bibitem[\protect\citeauthoryear{Gong, Eldele, Wu, Chen, Li, and Zhang}{Gong et~al\mbox{.}}{[n.d.]}]%
        {gongtowards}
\bibfield{author}{\bibinfo{person}{Peiliang Gong}, \bibinfo{person}{Emadeldeen Eldele}, \bibinfo{person}{Min Wu}, \bibinfo{person}{Zhenghua Chen}, \bibinfo{person}{Xiaoli Li}, {and} \bibinfo{person}{Daoqiang Zhang}.} \bibinfo{year}{[n.d.]}\natexlab{}.
\newblock \showarticletitle{Towards Adaptive Time Series Foundation Models Against Distribution Shift}.
\newblock  (\bibinfo{year}{[n.\,d.]}).
\newblock


\bibitem[\protect\citeauthoryear{Hasan, Islam, and Rahman}{Hasan et~al\mbox{.}}{2025}]%
        {hasan2025developing}
\bibfield{author}{\bibinfo{person}{MD~Rokibul Hasan}, \bibinfo{person}{Md~Raisul Islam}, {and} \bibinfo{person}{Md~Anisur Rahman}.} \bibinfo{year}{2025}\natexlab{}.
\newblock \showarticletitle{Developing and implementing AI-driven models for demand forecasting in US supply chains: A comprehensive approach to enhancing predictive accuracy}.
\newblock \bibinfo{journal}{\emph{Edelweiss applied science and technology}} \bibinfo{volume}{9}, \bibinfo{number}{1} (\bibinfo{year}{2025}), \bibinfo{pages}{1045--1068}.
\newblock


\bibitem[\protect\citeauthoryear{Hewamalage, Montero-Manso, Bergmeir, and Hyndman}{Hewamalage et~al\mbox{.}}{2021}]%
        {hewamalage2021look}
\bibfield{author}{\bibinfo{person}{Hansika Hewamalage}, \bibinfo{person}{Pablo Montero-Manso}, \bibinfo{person}{Christoph Bergmeir}, {and} \bibinfo{person}{Rob~J Hyndman}.} \bibinfo{year}{2021}\natexlab{}.
\newblock \showarticletitle{A look at the evaluation setup of the m5 forecasting competition}.
\newblock \bibinfo{journal}{\emph{arXiv preprint arXiv:2108.03588}} (\bibinfo{year}{2021}).
\newblock


\bibitem[\protect\citeauthoryear{Imambi, Prakash, and Kanagachidambaresan}{Imambi et~al\mbox{.}}{2021}]%
        {imambi2021pytorch}
\bibfield{author}{\bibinfo{person}{Sagar Imambi}, \bibinfo{person}{Kolla~Bhanu Prakash}, {and} \bibinfo{person}{GR Kanagachidambaresan}.} \bibinfo{year}{2021}\natexlab{}.
\newblock \showarticletitle{PyTorch}.
\newblock \bibinfo{journal}{\emph{Programming with TensorFlow: solution for edge computing applications}} (\bibinfo{year}{2021}), \bibinfo{pages}{87--104}.
\newblock


\bibitem[\protect\citeauthoryear{IN}{IN}{2020a}]%
        {kaggle-top1}
\bibfield{author}{\bibinfo{person}{Yeonjun IN}.} \bibinfo{year}{2020}\natexlab{a}.
\newblock \bibinfo{booktitle}{\emph{1st Place Solution: M5 Forecasting - Accuracy}}.
\newblock
\urldef\tempurl%
\url{https://github.com/YeonJun-IN/data.science.competition/blob/master/4.%20%5Bkaggle%5DM5-Accuracy/}
\showURL{%
\tempurl}
\newblock
\shownote{Accessed May 2025.}


\bibitem[\protect\citeauthoryear{IN}{IN}{2020b}]%
        {kaggle-top2}
\bibfield{author}{\bibinfo{person}{Yeonjun IN}.} \bibinfo{year}{2020}\natexlab{b}.
\newblock \bibinfo{booktitle}{\emph{1st Place Solution: M5 Forecasting - Accuracy}}.
\newblock
\urldef\tempurl%
\url{https://github.com/YeonJun-IN/data.science.competition/blob/master/4.%20%5Bkaggle%5DM5-Accuracy/}
\showURL{%
\tempurl}
\newblock
\shownote{Accessed May 2025.}


\bibitem[\protect\citeauthoryear{IN}{IN}{2020c}]%
        {kaggle-top3}
\bibfield{author}{\bibinfo{person}{Yeonjun IN}.} \bibinfo{year}{2020}\natexlab{c}.
\newblock \bibinfo{booktitle}{\emph{1st Place Solution: M5 Forecasting - Accuracy}}.
\newblock
\urldef\tempurl%
\url{https://github.com/YeonJun-IN/data.science.competition/blob/master/4.%20%5Bkaggle%5DM5-Accuracy/}
\showURL{%
\tempurl}
\newblock
\shownote{Accessed May 2025.}


\bibitem[\protect\citeauthoryear{Jahin, Shahriar, and Al~Amin}{Jahin et~al\mbox{.}}{2024}]%
        {jahin2024mcdfn}
\bibfield{author}{\bibinfo{person}{Md~Abrar Jahin}, \bibinfo{person}{Asef Shahriar}, {and} \bibinfo{person}{Md Al~Amin}.} \bibinfo{year}{2024}\natexlab{}.
\newblock \showarticletitle{Mcdfn: Supply chain demand forecasting via an explainable multi-channel data fusion network model integrating cnn, lstm, and gru}.
\newblock \bibinfo{journal}{\emph{arXiv e-prints}} (\bibinfo{year}{2024}), \bibinfo{pages}{arXiv--2405}.
\newblock


\bibitem[\protect\citeauthoryear{Jia, Wang, Zheng, Cao, and Liu}{Jia et~al\mbox{.}}{2024}]%
        {jia2024gpt4mts}
\bibfield{author}{\bibinfo{person}{Furong Jia}, \bibinfo{person}{Kevin Wang}, \bibinfo{person}{Yixiang Zheng}, \bibinfo{person}{Defu Cao}, {and} \bibinfo{person}{Yan Liu}.} \bibinfo{year}{2024}\natexlab{}.
\newblock \showarticletitle{Gpt4mts: Prompt-based large language model for multimodal time-series forecasting}. In \bibinfo{booktitle}{\emph{Proceedings of the AAAI Conference on Artificial Intelligence}}, Vol.~\bibinfo{volume}{38}. \bibinfo{pages}{23343--23351}.
\newblock


\bibitem[\protect\citeauthoryear{Jin, Wang, Ma, Chu, Zhang, Shi, Chen, Liang, Li, Pan, et~al\mbox{.}}{Jin et~al\mbox{.}}{2023}]%
        {jin2023time}
\bibfield{author}{\bibinfo{person}{Ming Jin}, \bibinfo{person}{Shiyu Wang}, \bibinfo{person}{Lintao Ma}, \bibinfo{person}{Zhixuan Chu}, \bibinfo{person}{James~Y Zhang}, \bibinfo{person}{Xiaoming Shi}, \bibinfo{person}{Pin-Yu Chen}, \bibinfo{person}{Yuxuan Liang}, \bibinfo{person}{Yuan-Fang Li}, \bibinfo{person}{Shirui Pan}, {et~al\mbox{.}}} \bibinfo{year}{2023}\natexlab{}.
\newblock \showarticletitle{Time-llm: Time series forecasting by reprogramming large language models}.
\newblock \bibinfo{journal}{\emph{arXiv preprint arXiv:2310.01728}} (\bibinfo{year}{2023}).
\newblock


\bibitem[\protect\citeauthoryear{Jin, Zhang, Chen, Zhang, Liang, Yang, Wang, Pan, and Wen}{Jin et~al\mbox{.}}{2024}]%
        {jin2024position}
\bibfield{author}{\bibinfo{person}{Ming Jin}, \bibinfo{person}{Yifan Zhang}, \bibinfo{person}{Wei Chen}, \bibinfo{person}{Kexin Zhang}, \bibinfo{person}{Yuxuan Liang}, \bibinfo{person}{Bin Yang}, \bibinfo{person}{Jindong Wang}, \bibinfo{person}{Shirui Pan}, {and} \bibinfo{person}{Qingsong Wen}.} \bibinfo{year}{2024}\natexlab{}.
\newblock \showarticletitle{Position Paper: What Can Large Language Models Tell Us about Time Series Analysis}.
\newblock \bibinfo{journal}{\emph{arXiv preprint arXiv:2402.02713}} (\bibinfo{year}{2024}).
\newblock


\bibitem[\protect\citeauthoryear{Kagalwala, Radhakrishnan, Mohammed, Kothinti, and Kulkarni}{Kagalwala et~al\mbox{.}}{2025}]%
        {kagalwala2025predictive}
\bibfield{author}{\bibinfo{person}{Hatim Kagalwala}, \bibinfo{person}{GV Radhakrishnan}, \bibinfo{person}{Irshadullah~Asim Mohammed}, \bibinfo{person}{Rishi~Reddy Kothinti}, {and} \bibinfo{person}{Nirzar Kulkarni}.} \bibinfo{year}{2025}\natexlab{}.
\newblock \showarticletitle{Predictive analytics in supply chain management: The role of AI and machine learning in demand forecasting}.
\newblock \bibinfo{journal}{\emph{Advances in Consumer Research}}  \bibinfo{volume}{2} (\bibinfo{year}{2025}), \bibinfo{pages}{142--149}.
\newblock


\bibitem[\protect\citeauthoryear{Ke, Meng, Finley, Wang, Chen, Ma, Ye, and Liu}{Ke et~al\mbox{.}}{2017}]%
        {ke2017lightgbm}
\bibfield{author}{\bibinfo{person}{Guolin Ke}, \bibinfo{person}{Qi Meng}, \bibinfo{person}{Thomas Finley}, \bibinfo{person}{Taifeng Wang}, \bibinfo{person}{Wei Chen}, \bibinfo{person}{Weidong Ma}, \bibinfo{person}{Qiwei Ye}, {and} \bibinfo{person}{Tie-Yan Liu}.} \bibinfo{year}{2017}\natexlab{}.
\newblock \showarticletitle{Lightgbm: A highly efficient gradient boosting decision tree}.
\newblock \bibinfo{journal}{\emph{Advances in neural information processing systems}}  \bibinfo{volume}{30} (\bibinfo{year}{2017}).
\newblock


\bibitem[\protect\citeauthoryear{Khlie, Benmamoun, Fethallah, and Jebbor}{Khlie et~al\mbox{.}}{2024}]%
        {khlie2024leveraging}
\bibfield{author}{\bibinfo{person}{Khaoula Khlie}, \bibinfo{person}{Z Benmamoun}, \bibinfo{person}{W Fethallah}, {and} \bibinfo{person}{I Jebbor}.} \bibinfo{year}{2024}\natexlab{}.
\newblock \showarticletitle{Leveraging variational autoencoders and recurrent neural networks for demand forecasting in supply chain management: A case study}.
\newblock \bibinfo{journal}{\emph{Journal of infrastructure, policy and development}} \bibinfo{volume}{8}, \bibinfo{number}{8} (\bibinfo{year}{2024}), \bibinfo{pages}{6639}.
\newblock


\bibitem[\protect\citeauthoryear{Kolassa}{Kolassa}{2022}]%
        {kolassa2022commentary}
\bibfield{author}{\bibinfo{person}{Stephan Kolassa}.} \bibinfo{year}{2022}\natexlab{}.
\newblock \showarticletitle{Commentary on the M5 forecasting competition}.
\newblock \bibinfo{journal}{\emph{International Journal of Forecasting}} \bibinfo{volume}{38}, \bibinfo{number}{4} (\bibinfo{year}{2022}), \bibinfo{pages}{1562--1568}.
\newblock


\bibitem[\protect\citeauthoryear{Kumar, Choubey, Amosu, and Ogunsuji}{Kumar et~al\mbox{.}}{2024}]%
        {kumar2024ai}
\bibfield{author}{\bibinfo{person}{Praveen Kumar}, \bibinfo{person}{Divya Choubey}, \bibinfo{person}{Olamide~Raimat Amosu}, {and} \bibinfo{person}{Yewande~Mariam Ogunsuji}.} \bibinfo{year}{2024}\natexlab{}.
\newblock \showarticletitle{AI-enhanced inventory and demand forecasting: Using AI to optimize inventory management and predict customer demand}.
\newblock \bibinfo{journal}{\emph{World J. Adv. Res. Rev}} \bibinfo{volume}{23}, \bibinfo{number}{1} (\bibinfo{year}{2024}).
\newblock


\bibitem[\protect\citeauthoryear{Lainder and Wolfinger}{Lainder and Wolfinger}{2022}]%
        {lainder2022forecasting}
\bibfield{author}{\bibinfo{person}{A~David Lainder} {and} \bibinfo{person}{Russell~D Wolfinger}.} \bibinfo{year}{2022}\natexlab{}.
\newblock \showarticletitle{Forecasting with gradient boosted trees: augmentation, tuning, and cross-validation strategies: Winning solution to the M5 Uncertainty competition}.
\newblock \bibinfo{journal}{\emph{International Journal of Forecasting}} \bibinfo{volume}{38}, \bibinfo{number}{4} (\bibinfo{year}{2022}), \bibinfo{pages}{1426--1433}.
\newblock


\bibitem[\protect\citeauthoryear{Li, Yang, Zhang, Xiao, Cao, Qin, Zhang, Zhao, and Bogdan}{Li et~al\mbox{.}}{2025}]%
        {li2025climatellm}
\bibfield{author}{\bibinfo{person}{Shixuan Li}, \bibinfo{person}{Wei Yang}, \bibinfo{person}{Peiyu Zhang}, \bibinfo{person}{Xiongye Xiao}, \bibinfo{person}{Defu Cao}, \bibinfo{person}{Yuehan Qin}, \bibinfo{person}{Xiaole Zhang}, \bibinfo{person}{Yue Zhao}, {and} \bibinfo{person}{Paul Bogdan}.} \bibinfo{year}{2025}\natexlab{}.
\newblock \showarticletitle{Climatellm: Efficient weather forecasting via frequency-aware large language models}.
\newblock \bibinfo{journal}{\emph{arXiv preprint arXiv:2502.11059}} (\bibinfo{year}{2025}).
\newblock


\bibitem[\protect\citeauthoryear{Li, Xu, Law, and Wang}{Li et~al\mbox{.}}{2024}]%
        {li2024enhancing}
\bibfield{author}{\bibinfo{person}{Xin Li}, \bibinfo{person}{Yechi Xu}, \bibinfo{person}{Rob Law}, {and} \bibinfo{person}{Shouyang Wang}.} \bibinfo{year}{2024}\natexlab{}.
\newblock \showarticletitle{Enhancing tourism demand forecasting with a transformer-based framework}.
\newblock \bibinfo{journal}{\emph{Annals of Tourism Research}}  \bibinfo{volume}{107} (\bibinfo{year}{2024}), \bibinfo{pages}{103791}.
\newblock


\bibitem[\protect\citeauthoryear{Liu, Li, Ji, Li, and Luo}{Liu et~al\mbox{.}}{2024b}]%
        {liu2024intelligent}
\bibfield{author}{\bibinfo{person}{Bojing Liu}, \bibinfo{person}{Mengxiang Li}, \bibinfo{person}{Zihui Ji}, \bibinfo{person}{Hongming Li}, {and} \bibinfo{person}{Ji Luo}.} \bibinfo{year}{2024}\natexlab{b}.
\newblock \showarticletitle{Intelligent productivity transformation: corporate market demand forecasting with the aid of an AI virtual assistant}.
\newblock \bibinfo{journal}{\emph{Journal of Organizational and End User Computing (JOEUC)}} \bibinfo{volume}{36}, \bibinfo{number}{1} (\bibinfo{year}{2024}), \bibinfo{pages}{1--27}.
\newblock


\bibitem[\protect\citeauthoryear{Liu, Xu, Miao, Yang, Zhang, Long, Li, and Zhao}{Liu et~al\mbox{.}}{2025}]%
        {liu2025timecma}
\bibfield{author}{\bibinfo{person}{Chenxi Liu}, \bibinfo{person}{Qianxiong Xu}, \bibinfo{person}{Hao Miao}, \bibinfo{person}{Sun Yang}, \bibinfo{person}{Lingzheng Zhang}, \bibinfo{person}{Cheng Long}, \bibinfo{person}{Ziyue Li}, {and} \bibinfo{person}{Rui Zhao}.} \bibinfo{year}{2025}\natexlab{}.
\newblock \showarticletitle{Timecma: Towards llm-empowered multivariate time series forecasting via cross-modality alignment}. In \bibinfo{booktitle}{\emph{Proceedings of the AAAI Conference on Artificial Intelligence}}, Vol.~\bibinfo{volume}{39}. \bibinfo{pages}{18780--18788}.
\newblock


\bibitem[\protect\citeauthoryear{Liu, Yu, Liao, Li, Lin, Liu, and Dustdar}{Liu et~al\mbox{.}}{2022}]%
        {liu2022pyraformer}
\bibfield{author}{\bibinfo{person}{Shizhan Liu}, \bibinfo{person}{Hang Yu}, \bibinfo{person}{Cong Liao}, \bibinfo{person}{Jianguo Li}, \bibinfo{person}{Weiyao Lin}, \bibinfo{person}{Alex~X Liu}, {and} \bibinfo{person}{Schahram Dustdar}.} \bibinfo{year}{2022}\natexlab{}.
\newblock \showarticletitle{Pyraformer: Low-complexity pyramidal attention for long-range time series modeling and forecasting}. In \bibinfo{booktitle}{\emph{\# PLACEHOLDER\_PARENT\_METADATA\_VALUE\#}}.
\newblock


\bibitem[\protect\citeauthoryear{Liu and Zhou}{Liu and Zhou}{2024}]%
        {liu2024using}
\bibfield{author}{\bibinfo{person}{Siwei Liu} {and} \bibinfo{person}{Di~Jody Zhou}.} \bibinfo{year}{2024}\natexlab{}.
\newblock \showarticletitle{Using cross-validation methods to select time series models: Promises and pitfalls}.
\newblock \bibinfo{journal}{\emph{Brit. J. Math. Statist. Psych.}} \bibinfo{volume}{77}, \bibinfo{number}{2} (\bibinfo{year}{2024}), \bibinfo{pages}{337--355}.
\newblock


\bibitem[\protect\citeauthoryear{Liu, Hu, Li, Diao, Liang, Hooi, and Zimmermann}{Liu et~al\mbox{.}}{2024a}]%
        {liu2024unitime}
\bibfield{author}{\bibinfo{person}{Xu Liu}, \bibinfo{person}{Junfeng Hu}, \bibinfo{person}{Yuan Li}, \bibinfo{person}{Shizhe Diao}, \bibinfo{person}{Yuxuan Liang}, \bibinfo{person}{Bryan Hooi}, {and} \bibinfo{person}{Roger Zimmermann}.} \bibinfo{year}{2024}\natexlab{a}.
\newblock \showarticletitle{Unitime: A language-empowered unified model for cross-domain time series forecasting}. In \bibinfo{booktitle}{\emph{Proceedings of the ACM Web Conference 2024}}. \bibinfo{pages}{4095--4106}.
\newblock


\bibitem[\protect\citeauthoryear{Ma, Liu, Zheng, Huang, Zhu, Yu, and Kwok}{Ma et~al\mbox{.}}{2024}]%
        {ma2024survey}
\bibfield{author}{\bibinfo{person}{Qianli Ma}, \bibinfo{person}{Zhen Liu}, \bibinfo{person}{Zhenjing Zheng}, \bibinfo{person}{Ziyang Huang}, \bibinfo{person}{Siying Zhu}, \bibinfo{person}{Zhongzhong Yu}, {and} \bibinfo{person}{James~T Kwok}.} \bibinfo{year}{2024}\natexlab{}.
\newblock \showarticletitle{A survey on time-series pre-trained models}.
\newblock \bibinfo{journal}{\emph{IEEE Transactions on Knowledge and Data Engineering}} (\bibinfo{year}{2024}).
\newblock


\bibitem[\protect\citeauthoryear{Makridakis, Petropoulos, and Spiliotis}{Makridakis et~al\mbox{.}}{2022}]%
        {makridakis2022m5}
\bibfield{author}{\bibinfo{person}{Spyros Makridakis}, \bibinfo{person}{Fotios Petropoulos}, {and} \bibinfo{person}{Evangelos Spiliotis}.} \bibinfo{year}{2022}\natexlab{}.
\newblock \bibinfo{title}{The M5 competition: conclusions}.
\newblock , \bibinfo{numpages}{1576--1582}~pages.
\newblock


\bibitem[\protect\citeauthoryear{Mitra, Jain, Kishore, and Kumar}{Mitra et~al\mbox{.}}{2022}]%
        {mitra2022comparative}
\bibfield{author}{\bibinfo{person}{Arnab Mitra}, \bibinfo{person}{Arnav Jain}, \bibinfo{person}{Avinash Kishore}, {and} \bibinfo{person}{Pravin Kumar}.} \bibinfo{year}{2022}\natexlab{}.
\newblock \showarticletitle{A comparative study of demand forecasting models for a multi-channel retail company: a novel hybrid machine learning approach}. In \bibinfo{booktitle}{\emph{Operations research forum}}, Vol.~\bibinfo{volume}{3}. Springer, \bibinfo{pages}{58}.
\newblock


\bibitem[\protect\citeauthoryear{Mu{\~n}oz-Zavala, Mac{\'\i}as-D{\'\i}az, Alba-Cu{\'e}llar, and Guerrero-D{\'\i}az-de Le{\'o}n}{Mu{\~n}oz-Zavala et~al\mbox{.}}{2024}]%
        {munoz2024literature}
\bibfield{author}{\bibinfo{person}{Angel~E Mu{\~n}oz-Zavala}, \bibinfo{person}{Jorge~E Mac{\'\i}as-D{\'\i}az}, \bibinfo{person}{Daniel Alba-Cu{\'e}llar}, {and} \bibinfo{person}{Jos{\'e}~A Guerrero-D{\'\i}az-de Le{\'o}n}.} \bibinfo{year}{2024}\natexlab{}.
\newblock \showarticletitle{A literature review on some trends in artificial neural networks for modeling and simulation with time series}.
\newblock \bibinfo{journal}{\emph{Algorithms}} \bibinfo{volume}{17}, \bibinfo{number}{2} (\bibinfo{year}{2024}), \bibinfo{pages}{76}.
\newblock


\bibitem[\protect\citeauthoryear{Nia, Awasthi, and Bhuiyan}{Nia et~al\mbox{.}}{2021}]%
        {nia2021industry}
\bibfield{author}{\bibinfo{person}{Ali~Roozbeh Nia}, \bibinfo{person}{Anjali Awasthi}, {and} \bibinfo{person}{Nadia Bhuiyan}.} \bibinfo{year}{2021}\natexlab{}.
\newblock \showarticletitle{Industry 4.0 and demand forecasting of the energy supply chain: A literature review}.
\newblock \bibinfo{journal}{\emph{Computers \& Industrial Engineering}}  \bibinfo{volume}{154} (\bibinfo{year}{2021}), \bibinfo{pages}{107128}.
\newblock


\bibitem[\protect\citeauthoryear{Nie, Nguyen, Sinthong, and Kalagnanam}{Nie et~al\mbox{.}}{2022}]%
        {nie2022time}
\bibfield{author}{\bibinfo{person}{Yuqi Nie}, \bibinfo{person}{Nam~H Nguyen}, \bibinfo{person}{Phanwadee Sinthong}, {and} \bibinfo{person}{Jayant Kalagnanam}.} \bibinfo{year}{2022}\natexlab{}.
\newblock \showarticletitle{A time series is worth 64 words: Long-term forecasting with transformers}.
\newblock \bibinfo{journal}{\emph{arXiv preprint arXiv:2211.14730}} (\bibinfo{year}{2022}).
\newblock


\bibitem[\protect\citeauthoryear{Niu, Habault, Cao, Zhang, Legaspi, Ung, Enouen, Wada, Ono, Minamikawa, et~al\mbox{.}}{Niu et~al\mbox{.}}{2024}]%
        {niu2024mixture}
\bibfield{author}{\bibinfo{person}{Hao Niu}, \bibinfo{person}{Guillaume Habault}, \bibinfo{person}{Defu Cao}, \bibinfo{person}{Yizhou Zhang}, \bibinfo{person}{Roberto Legaspi}, \bibinfo{person}{Huy~Quang Ung}, \bibinfo{person}{James Enouen}, \bibinfo{person}{Shinya Wada}, \bibinfo{person}{Chihiro Ono}, \bibinfo{person}{Atsunori Minamikawa}, {et~al\mbox{.}}} \bibinfo{year}{2024}\natexlab{}.
\newblock \showarticletitle{Mixture of Projection Experts for Multivariate Long-Term Time Series Forecasting}. In \bibinfo{booktitle}{\emph{2024 International Conference on Machine Learning and Applications (ICMLA)}}. IEEE, \bibinfo{pages}{1798--1803}.
\newblock


\bibitem[\protect\citeauthoryear{Niu, Habault, Legaspi, Meng, Cao, Wada, Ono, and Liu}{Niu et~al\mbox{.}}{2023}]%
        {niu2023time}
\bibfield{author}{\bibinfo{person}{Hao Niu}, \bibinfo{person}{Guillaume Habault}, \bibinfo{person}{Roberto Legaspi}, \bibinfo{person}{Chuizheng Meng}, \bibinfo{person}{Defu Cao}, \bibinfo{person}{Shinya Wada}, \bibinfo{person}{Chihiro Ono}, {and} \bibinfo{person}{Yan Liu}.} \bibinfo{year}{2023}\natexlab{}.
\newblock \showarticletitle{Time-delayed multivariate time series predictions}. In \bibinfo{booktitle}{\emph{Proceedings of the 2023 SIAM International Conference on Data Mining (SDM)}}. SIAM, \bibinfo{pages}{325--333}.
\newblock


\bibitem[\protect\citeauthoryear{Oyewole, Okoye, Ofodile, and Ejairu}{Oyewole et~al\mbox{.}}{2024}]%
        {oyewole2024reviewing}
\bibfield{author}{\bibinfo{person}{Adedoyin~Tolulope Oyewole}, \bibinfo{person}{Chinwe~Chinazo Okoye}, \bibinfo{person}{Onyeka~Chrisanctus Ofodile}, {and} \bibinfo{person}{Emuesiri Ejairu}.} \bibinfo{year}{2024}\natexlab{}.
\newblock \showarticletitle{Reviewing predictive analytics in supply chain management: Applications and benefits}.
\newblock \bibinfo{journal}{\emph{World Journal of Advanced Research and Reviews}} \bibinfo{volume}{21}, \bibinfo{number}{3} (\bibinfo{year}{2024}), \bibinfo{pages}{568--574}.
\newblock


\bibitem[\protect\citeauthoryear{Rasul, Ashok, Williams, Khorasani, Adamopoulos, Bhagwatkar, Bilo{\v{s}}, Ghonia, Hassen, Schneider, et~al\mbox{.}}{Rasul et~al\mbox{.}}{2023}]%
        {rasul2023lag}
\bibfield{author}{\bibinfo{person}{Kashif Rasul}, \bibinfo{person}{Arjun Ashok}, \bibinfo{person}{Andrew~Robert Williams}, \bibinfo{person}{Arian Khorasani}, \bibinfo{person}{George Adamopoulos}, \bibinfo{person}{Rishika Bhagwatkar}, \bibinfo{person}{Marin Bilo{\v{s}}}, \bibinfo{person}{Hena Ghonia}, \bibinfo{person}{Nadhir~Vincent Hassen}, \bibinfo{person}{Anderson Schneider}, {et~al\mbox{.}}} \bibinfo{year}{2023}\natexlab{}.
\newblock \showarticletitle{Lag-llama: Towards foundation models for time series forecasting}.
\newblock \bibinfo{journal}{\emph{arXiv preprint arXiv:2310.08278}} (\bibinfo{year}{2023}).
\newblock


\bibitem[\protect\citeauthoryear{Salinas, Flunkert, Gasthaus, and Januschowski}{Salinas et~al\mbox{.}}{2020}]%
        {salinas2020deepar}
\bibfield{author}{\bibinfo{person}{David Salinas}, \bibinfo{person}{Valentin Flunkert}, \bibinfo{person}{Jan Gasthaus}, {and} \bibinfo{person}{Tim Januschowski}.} \bibinfo{year}{2020}\natexlab{}.
\newblock \showarticletitle{DeepAR: Probabilistic forecasting with autoregressive recurrent networks}.
\newblock \bibinfo{journal}{\emph{International journal of forecasting}} \bibinfo{volume}{36}, \bibinfo{number}{3} (\bibinfo{year}{2020}), \bibinfo{pages}{1181--1191}.
\newblock


\bibitem[\protect\citeauthoryear{Seaman and Bowman}{Seaman and Bowman}{2022}]%
        {seaman2022applicability}
\bibfield{author}{\bibinfo{person}{Brian Seaman} {and} \bibinfo{person}{John Bowman}.} \bibinfo{year}{2022}\natexlab{}.
\newblock \showarticletitle{Applicability of the M5 to Forecasting at Walmart}.
\newblock \bibinfo{journal}{\emph{International Journal of Forecasting}} \bibinfo{volume}{38}, \bibinfo{number}{4} (\bibinfo{year}{2022}), \bibinfo{pages}{1468--1472}.
\newblock


\bibitem[\protect\citeauthoryear{Shi, Wang, Nie, Li, Ye, Wen, and Jin}{Shi et~al\mbox{.}}{2024}]%
        {shi2024time}
\bibfield{author}{\bibinfo{person}{Xiaoming Shi}, \bibinfo{person}{Shiyu Wang}, \bibinfo{person}{Yuqi Nie}, \bibinfo{person}{Dianqi Li}, \bibinfo{person}{Zhou Ye}, \bibinfo{person}{Qingsong Wen}, {and} \bibinfo{person}{Ming Jin}.} \bibinfo{year}{2024}\natexlab{}.
\newblock \showarticletitle{Time-moe: Billion-scale time series foundation models with mixture of experts}.
\newblock \bibinfo{journal}{\emph{arXiv preprint arXiv:2409.16040}} (\bibinfo{year}{2024}).
\newblock


\bibitem[\protect\citeauthoryear{Spiliotis, Makridakis, Kaltsounis, and Assimakopoulos}{Spiliotis et~al\mbox{.}}{2021}]%
        {spiliotis2021product}
\bibfield{author}{\bibinfo{person}{Evangelos Spiliotis}, \bibinfo{person}{Spyros Makridakis}, \bibinfo{person}{Anastasios Kaltsounis}, {and} \bibinfo{person}{Vassilios Assimakopoulos}.} \bibinfo{year}{2021}\natexlab{}.
\newblock \showarticletitle{Product sales probabilistic forecasting: An empirical evaluation using the M5 competition data}.
\newblock \bibinfo{journal}{\emph{International Journal of Production Economics}}  \bibinfo{volume}{240} (\bibinfo{year}{2021}), \bibinfo{pages}{108237}.
\newblock


\bibitem[\protect\citeauthoryear{Swaminathan and Venkitasubramony}{Swaminathan and Venkitasubramony}{2024}]%
        {swaminathan2024demand}
\bibfield{author}{\bibinfo{person}{Kritika Swaminathan} {and} \bibinfo{person}{Rakesh Venkitasubramony}.} \bibinfo{year}{2024}\natexlab{}.
\newblock \showarticletitle{Demand forecasting for fashion products: A systematic review}.
\newblock \bibinfo{journal}{\emph{International Journal of Forecasting}} \bibinfo{volume}{40}, \bibinfo{number}{1} (\bibinfo{year}{2024}), \bibinfo{pages}{247--267}.
\newblock


\bibitem[\protect\citeauthoryear{Sze, Chen, Yang, and Emer}{Sze et~al\mbox{.}}{2017}]%
        {sze2017efficient}
\bibfield{author}{\bibinfo{person}{Vivienne Sze}, \bibinfo{person}{Yu-Hsin Chen}, \bibinfo{person}{Tien-Ju Yang}, {and} \bibinfo{person}{Joel~S Emer}.} \bibinfo{year}{2017}\natexlab{}.
\newblock \showarticletitle{Efficient processing of deep neural networks: A tutorial and survey}.
\newblock \bibinfo{journal}{\emph{Proc. IEEE}} \bibinfo{volume}{105}, \bibinfo{number}{12} (\bibinfo{year}{2017}), \bibinfo{pages}{2295--2329}.
\newblock


\bibitem[\protect\citeauthoryear{Theodorou, Wang, Kang, Spiliotis, Makridakis, and Assimakopoulos}{Theodorou et~al\mbox{.}}{2022}]%
        {theodorou2022exploring}
\bibfield{author}{\bibinfo{person}{Evangelos Theodorou}, \bibinfo{person}{Shengjie Wang}, \bibinfo{person}{Yanfei Kang}, \bibinfo{person}{Evangelos Spiliotis}, \bibinfo{person}{Spyros Makridakis}, {and} \bibinfo{person}{Vassilios Assimakopoulos}.} \bibinfo{year}{2022}\natexlab{}.
\newblock \showarticletitle{Exploring the representativeness of the M5 competition data}.
\newblock \bibinfo{journal}{\emph{International Journal of Forecasting}} \bibinfo{volume}{38}, \bibinfo{number}{4} (\bibinfo{year}{2022}), \bibinfo{pages}{1500--1506}.
\newblock


\bibitem[\protect\citeauthoryear{Trapero, de~Frutos, and Pedregal}{Trapero et~al\mbox{.}}{2024}]%
        {trapero2024demand}
\bibfield{author}{\bibinfo{person}{Juan~R Trapero}, \bibinfo{person}{Enrique~Holgado de Frutos}, {and} \bibinfo{person}{Diego~J Pedregal}.} \bibinfo{year}{2024}\natexlab{}.
\newblock \showarticletitle{Demand forecasting under lost sales stock policies}.
\newblock \bibinfo{journal}{\emph{International Journal of Forecasting}} \bibinfo{volume}{40}, \bibinfo{number}{3} (\bibinfo{year}{2024}), \bibinfo{pages}{1055--1068}.
\newblock


\bibitem[\protect\citeauthoryear{Vallarino}{Vallarino}{2024}]%
        {vallarino2024dynamic}
\bibfield{author}{\bibinfo{person}{Diego Vallarino}.} \bibinfo{year}{2024}\natexlab{}.
\newblock \showarticletitle{A Dynamic Approach to Stock Price Prediction: Comparing RNN and Mixture of Experts Models Across Different Volatility Profiles}.
\newblock \bibinfo{journal}{\emph{arXiv preprint arXiv:2410.07234}} (\bibinfo{year}{2024}).
\newblock


\bibitem[\protect\citeauthoryear{Verma}{Verma}{2024}]%
        {verma2024transforming}
\bibfield{author}{\bibinfo{person}{Pradeep Verma}.} \bibinfo{year}{2024}\natexlab{}.
\newblock \showarticletitle{Transforming Supply Chains Through AI: Demand Forecasting, Inventory Management, and Dynamic Optimization}.
\newblock \bibinfo{journal}{\emph{Integrated Journal of Science and Technology}} \bibinfo{volume}{1}, \bibinfo{number}{9} (\bibinfo{year}{2024}).
\newblock


\bibitem[\protect\citeauthoryear{Wellens, Udenio, and Boute}{Wellens et~al\mbox{.}}{2022}]%
        {wellens2022transfer}
\bibfield{author}{\bibinfo{person}{Arnoud~P Wellens}, \bibinfo{person}{Maxi Udenio}, {and} \bibinfo{person}{Robert~N Boute}.} \bibinfo{year}{2022}\natexlab{}.
\newblock \showarticletitle{Transfer learning for hierarchical forecasting: Reducing computational efforts of M5 winning methods}.
\newblock \bibinfo{journal}{\emph{International Journal of Forecasting}} \bibinfo{volume}{38}, \bibinfo{number}{4} (\bibinfo{year}{2022}), \bibinfo{pages}{1482--1491}.
\newblock


\bibitem[\protect\citeauthoryear{Woo, Liu, Kumar, Xiong, Savarese, and Sahoo}{Woo et~al\mbox{.}}{2024}]%
        {woo2024unified}
\bibfield{author}{\bibinfo{person}{Gerald Woo}, \bibinfo{person}{Chenghao Liu}, \bibinfo{person}{Akshat Kumar}, \bibinfo{person}{Caiming Xiong}, \bibinfo{person}{Silvio Savarese}, {and} \bibinfo{person}{Doyen Sahoo}.} \bibinfo{year}{2024}\natexlab{}.
\newblock \showarticletitle{Unified training of universal time series forecasting transformers}.
\newblock  (\bibinfo{year}{2024}).
\newblock


\bibitem[\protect\citeauthoryear{Ye, Yang, Cao, Zhang, Tang, Cai, and Liu}{Ye et~al\mbox{.}}{2024}]%
        {ye2024domain}
\bibfield{author}{\bibinfo{person}{Wen Ye}, \bibinfo{person}{Wei Yang}, \bibinfo{person}{Defu Cao}, \bibinfo{person}{Yizhou Zhang}, \bibinfo{person}{Lumingyuan Tang}, \bibinfo{person}{Jie Cai}, {and} \bibinfo{person}{Yan Liu}.} \bibinfo{year}{2024}\natexlab{}.
\newblock \showarticletitle{Domain-Oriented Time Series Inference Agents for Reasoning and Automated Analysis}.
\newblock \bibinfo{journal}{\emph{arXiv preprint arXiv:2410.04047}} (\bibinfo{year}{2024}).
\newblock


\bibitem[\protect\citeauthoryear{Zhang, Li, Han, Yang, and Cui}{Zhang et~al\mbox{.}}{2024}]%
        {zhang2024enhancing}
\bibfield{author}{\bibinfo{person}{Xuguang Zhang}, \bibinfo{person}{Pan Li}, \bibinfo{person}{Xu Han}, \bibinfo{person}{Yongbin Yang}, {and} \bibinfo{person}{Yiwen Cui}.} \bibinfo{year}{2024}\natexlab{}.
\newblock \showarticletitle{Enhancing Time Series Product Demand Forecasting with Hybrid Attention-Based Deep Learning Models}.
\newblock \bibinfo{journal}{\emph{IEEE Access}} (\bibinfo{year}{2024}).
\newblock


\bibitem[\protect\citeauthoryear{Zhang and Yan}{Zhang and Yan}{2023}]%
        {zhang2023crossformer}
\bibfield{author}{\bibinfo{person}{Yunhao Zhang} {and} \bibinfo{person}{Junchi Yan}.} \bibinfo{year}{2023}\natexlab{}.
\newblock \showarticletitle{Crossformer: Transformer utilizing cross-dimension dependency for multivariate time series forecasting}. In \bibinfo{booktitle}{\emph{The eleventh international conference on learning representations}}.
\newblock


\bibitem[\protect\citeauthoryear{Zhou, Zhang, Peng, Zhang, Li, Xiong, and Zhang}{Zhou et~al\mbox{.}}{2021}]%
        {zhou2021informer}
\bibfield{author}{\bibinfo{person}{Haoyi Zhou}, \bibinfo{person}{Shanghang Zhang}, \bibinfo{person}{Jieqi Peng}, \bibinfo{person}{Shuai Zhang}, \bibinfo{person}{Jianxin Li}, \bibinfo{person}{Hui Xiong}, {and} \bibinfo{person}{Wancai Zhang}.} \bibinfo{year}{2021}\natexlab{}.
\newblock \showarticletitle{Informer: Beyond efficient transformer for long sequence time-series forecasting}. In \bibinfo{booktitle}{\emph{Proceedings of the AAAI conference on artificial intelligence}}, Vol.~\bibinfo{volume}{35}. \bibinfo{pages}{11106--11115}.
\newblock


\bibitem[\protect\citeauthoryear{Zhou, Niu, Sun, Jin, et~al\mbox{.}}{Zhou et~al\mbox{.}}{2023}]%
        {zhou2023one}
\bibfield{author}{\bibinfo{person}{Tian Zhou}, \bibinfo{person}{Peisong Niu}, \bibinfo{person}{Liang Sun}, \bibinfo{person}{Rong Jin}, {et~al\mbox{.}}} \bibinfo{year}{2023}\natexlab{}.
\newblock \showarticletitle{One fits all: Power general time series analysis by pretrained lm}.
\newblock \bibinfo{journal}{\emph{Advances in neural information processing systems}}  \bibinfo{volume}{36} (\bibinfo{year}{2023}), \bibinfo{pages}{43322--43355}.
\newblock


\bibitem[\protect\citeauthoryear{Ziel}{Ziel}{2022}]%
        {ziel2022m5}
\bibfield{author}{\bibinfo{person}{Florian Ziel}.} \bibinfo{year}{2022}\natexlab{}.
\newblock \showarticletitle{M5 competition uncertainty: Overdispersion, distributional forecasting, GAMLSS, and beyond}.
\newblock \bibinfo{journal}{\emph{International Journal of Forecasting}} \bibinfo{volume}{38}, \bibinfo{number}{4} (\bibinfo{year}{2022}), \bibinfo{pages}{1546--1554}.
\newblock


\end{thebibliography}

\end{document}